\documentclass[10pt,twocolumn,letterpaper]{article}

\usepackage[pagenumbers]{cvpr} 

\usepackage{graphicx}
\usepackage{amsmath}
\usepackage{amssymb}
\usepackage{booktabs}
\usepackage{xfrac}

\usepackage{tabulary,multirow}

\newcolumntype{x}[1]{>{\centering\arraybackslash}p{#1pt}}
\newlength\savewidth
\newcommand{\tablestyle}[2]{\setlength{\tabcolsep}{#1}\renewcommand{\arraystretch}{#2}\centering\footnotesize}
\makeatletter\renewcommand\paragraph{\@startsection{paragraph}{4}{\z@}
  {.5em \@plus.2ex \@minus.2ex}{-.5em}{\normalfont\normalsize\bfseries}}\makeatother
\usepackage[pagebackref,breaklinks,colorlinks]{hyperref}

\usepackage[capitalize]{cleveref}
\crefname{section}{Sec.}{Secs.}
\Crefname{section}{Section}{Sections}
\Crefname{table}{Table}{Tables}
\crefname{table}{Tab.}{Tabs.}

\def\cvprPaperID{} 
\def\confName{x}
\def\confYear{2021}

\begin{document}

\title{Q-ViT: Fully Differentiable Quantization for Vision Transformer\thanks{The work is done during Zhexin Li's internship at Megvii Technology. The code is released at \url{https://github.com/zhexinli/Q-ViT-DeiT}.\\
\quad $\dagger$ Corresponding author.}}
\author{Zhexin Li$^{1,3}$
\and
Tong Yang$^2$
\and
Peisong Wang$^1$
\and
Jian Cheng$^{1,\dagger}$
\and
$^1$Institute of Automation, Chinese Academy of Sciences $^2$Megvii Technology \\
$^3$School of Artificial Intelligence, University of Chinese Academy of Sciences\\
{\tt\small \{zhexin.li, peisong.wang, jcheng\}@nlpr.ia.ac.cn, yangtong@megvii.com}
}
\maketitle

\begin{abstract}
In this paper, we propose a fully differentiable quantization method for vision transformer (ViT) named as Q-ViT, in which both of the quantization scales and bit-widths are learnable parameters. Specifically, based on our observation that heads in ViT display different quantization robustness, we leverage head-wise bit-width to squeeze the size of Q-ViT while preserving performance. In addition, we propose a novel technique named switchable scale to resolve the convergence problem in the joint training of quantization scales and bit-widths. In this way, Q-ViT pushes the limits of ViT quantization to 3-bit without heavy performance drop. Moreover, we analyze the quantization robustness of every architecture component of ViT and show that the Multi-head Self-Attention (MSA) and the Gaussian Error Linear Units (GELU) are the key aspects for ViT quantization. This study provides some insights for further research about ViT quantization. Extensive experiments on different ViT models, such as DeiT and Swin Transformer show the effectiveness of our quantization method. In particular, our method outperforms the state-of-the-art uniform quantization method by 1.5\% on DeiT-Tiny.
\end{abstract}

\section{Introduction}
\label{sec:intro}

\begin{figure}[t]
    \centering
    \includegraphics[width=0.8\linewidth]{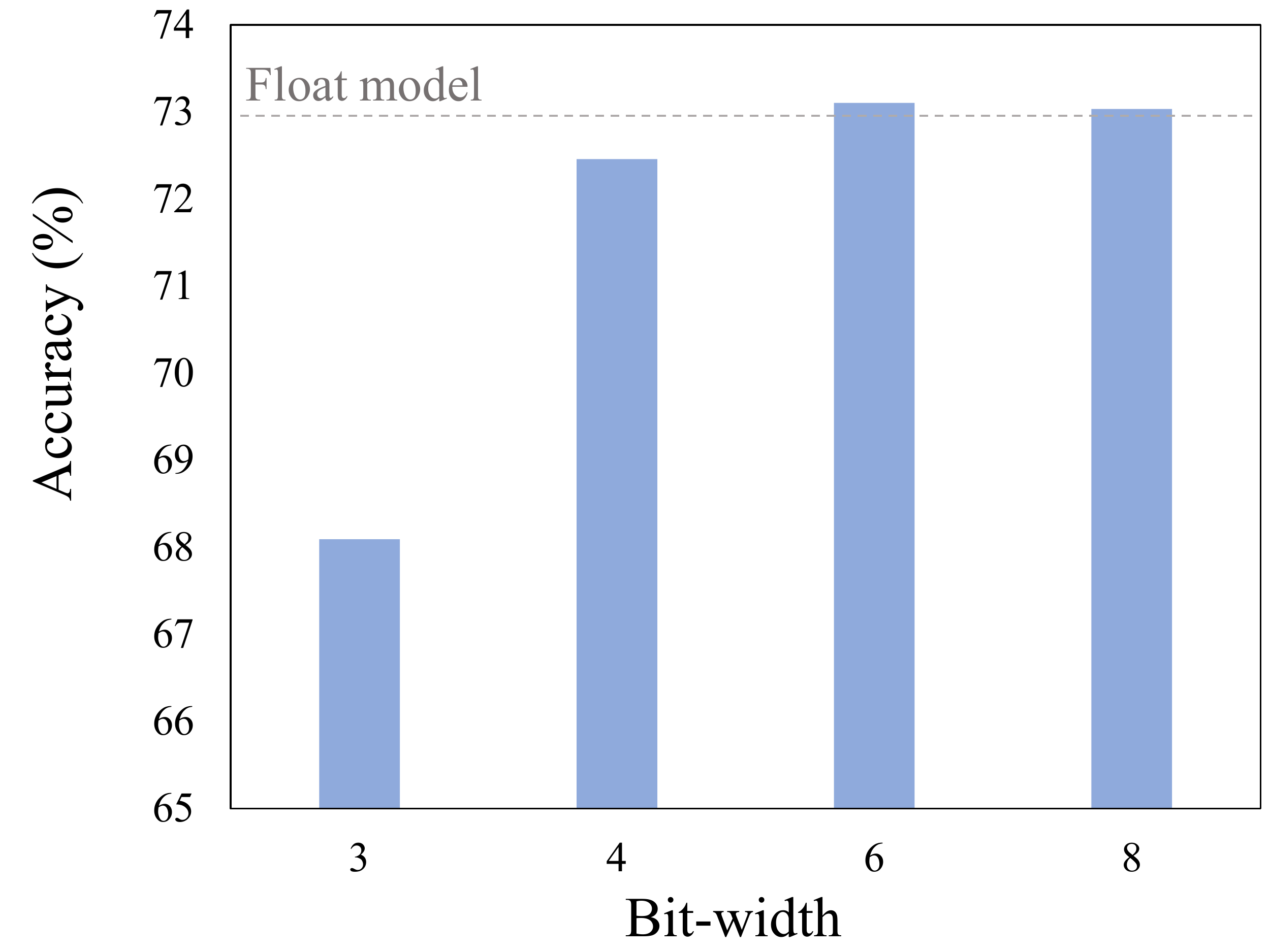}
    \caption{Quantization results for DeiT-Tiny on ImageNet, using a state-of-the-art QAT approach LSQ+. The accuracy encounters severe degradation when the bit-width decreases to 3-bit.}
    \label{fig:qat-limit}
\end{figure}
        
Recently, the vision transformer (ViT) has achieved great success in computer vision, such as image classification \cite{deit, vit, swin},  object detection \cite{detr}, and instance segmentation \cite{swin}. However, the vision transformer suffers from heavy computation and intensive memory cost. Thus, it is impractical to apply large vision transformer models to edge devices. To resolve these problems, many compression technologies, such as structured pruning \cite{iared, tokenlearner, dynamicvit}, quantization \cite{vitptq}, linear transformer \cite{combiner}, and customized acceleration \cite{reuse}, are applied to vision transformer. Although quantization has been massively studied for Convolution Neural Networks (CNN), few attention has been drawn to the quantization of ViT.
\begin{figure*}
    \centering
    \begin{subfigure}{0.5\linewidth}
        \includegraphics[width=1\linewidth]{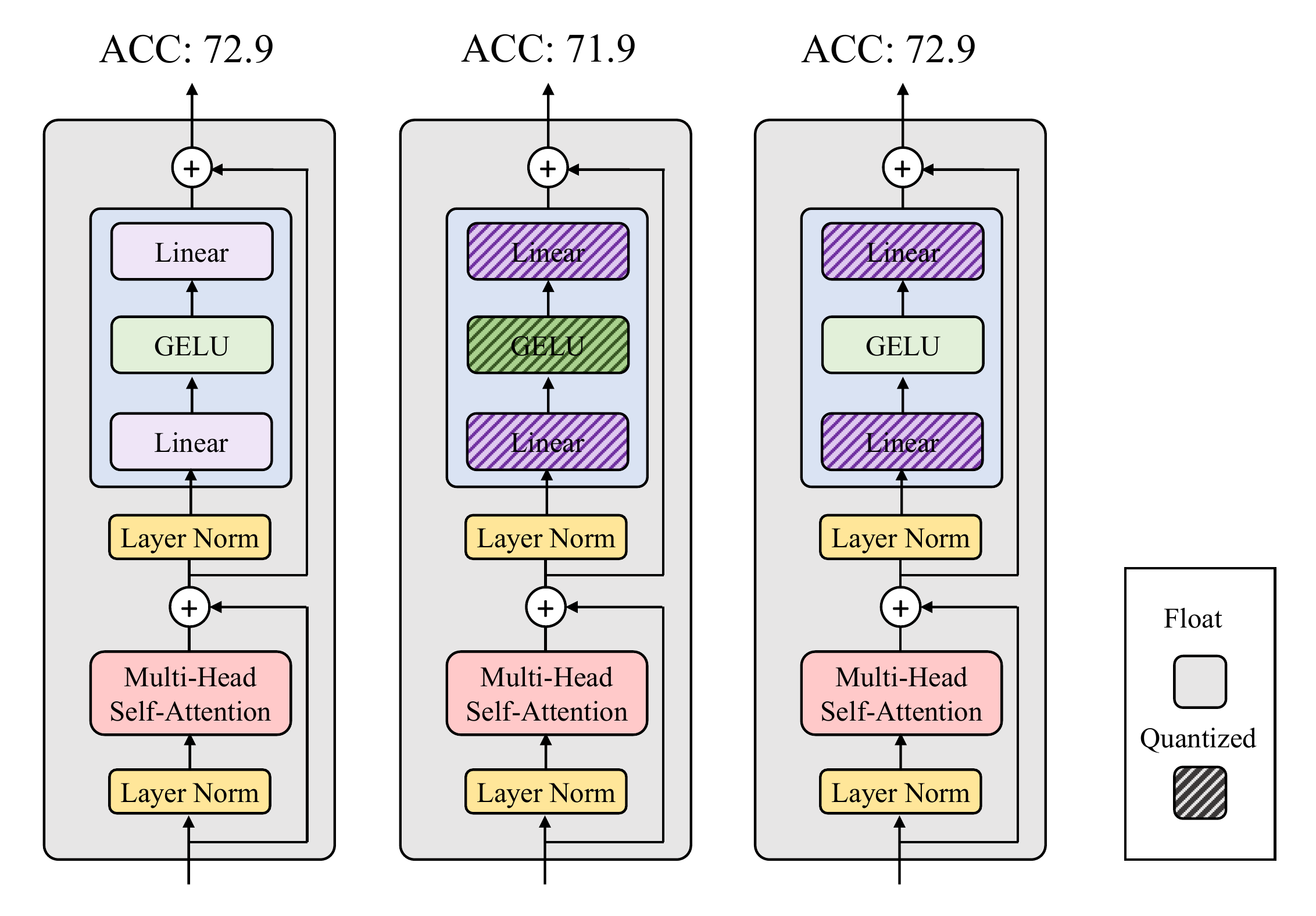}
        \caption{}
        \label{fig:sensitivity-a}
    \end{subfigure}
    \begin{subfigure}{0.43\linewidth}
        \includegraphics[width=1\linewidth]{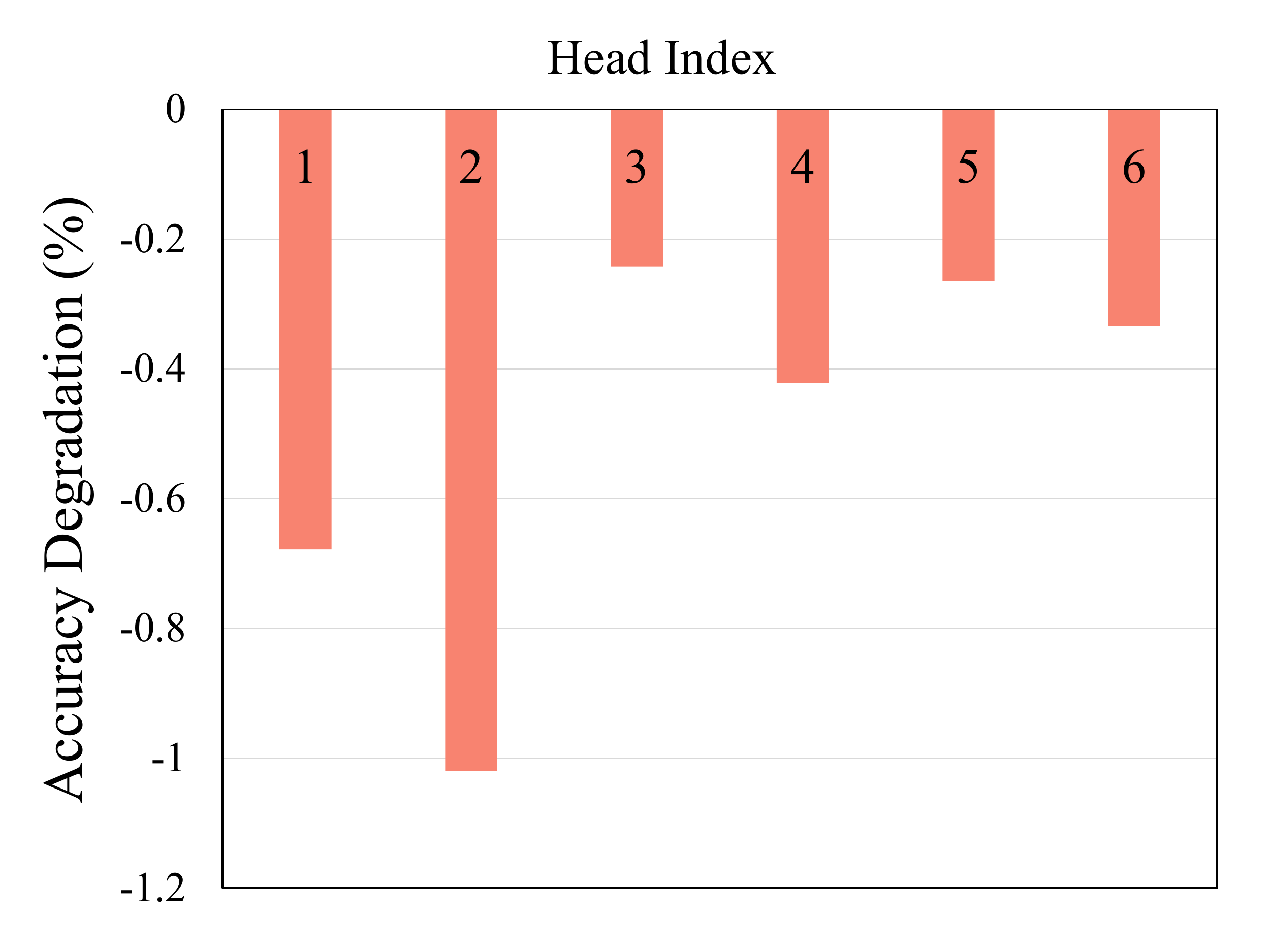}
        \caption{}
        \label{fig:sensitivity-b}
    \end{subfigure}
    \caption{Analysis of robustness against quantization noise for different structures in ViT. (a) Robustness analysis for the MLP layers in DeiT-Tiny, finetuned for 100 epochs using uniform quantization. The GELU activation layer shows dominant quantization error in MLP. (b) Head-wise quantization robustness analysis for a 8-bit quantized DeiT-Small. The table shows the accuracy degradation when quantizing a specific attention head in the first MSA layer to 2-bit. Heads of index 1 and 2 are significant. Allocating low bits to them causes severe performance degradation. The other heads can be quantized to 2-bit representations with mild performance drop.}
    \label{fig:sensitivity}
\end{figure*}
For quantization, there are two approaches: Post-Training Quantization (PTQ) and Quantization-Aware Training (QAT). Although PTQ is efficient for deployment, it is hard to achieve acceptable performance in ViT. For example, \cite{vitptq} only pushes its limit of quantization to 6-bit without great performance degradation. In contrast, QAT simulates quantization noise during the finetuning stage and shows more robustness to low-bit quantization. Results on CNN-based models have shown QAT is able to achieve low-bit quantization with negligible performance drop. Considering performance and robustness, it is potential to use QAT for low-bit ViT quantization.

As shown in \cref{fig:qat-limit}, existing QAT methods suffer from a great performance drop when the ViT is quantized to 3- or 4- bits of precision. This is different from LSQ \cite{lsq}, in which the low-bit quantized CNN can reach full-precision accuracy. For deeper understanding, we conduct experiments to explore the quantization robustness of different components in the transformer layers. As is shown in \cref{fig:sensitivity-a}, in MLP, the GELU \cite{gelu} is sensitive to quantization and need to be assigned with quantization bit-widths carefully.
In addition, in MSA, the quantization robustness of different heads varies very differently, see \cref{fig:sensitivity-b}. Based on these observations, we conclude that it is a must to design a special mixed-precision QAT for low-bit ViT quantization. 

In this paper, we propose a fully differentiable quantization for vision transformer, named as Q-ViT. Inspired by previous differentiable methods \cite{lsq, lsq+, dq, fracbits, edmips}, Q-ViT learns quantization scales and bit-widths for all components in ViT under the supervision of the classification loss and computation constraint. Specifically, based on the quantization robustness analysis, we propose head-wise bit-width allocation scheme to allow heads in MSA to learn different bit-widths according to their quantization robustness. The learned bit-width allocation scheme in Q-ViT also assigns GELU \cite{gelu} activation layers with high bits due to its sensitiveness to quantization. In addition, we also propose a novel technique, named switchable scale, to solve the instability problem emerging when scales and bit-widths are learned simultaneously.

Extensive experiments are conducted to show the effectiveness of our method on different ViT models, such as DeiT \cite{deit} and Swin Transformer \cite{swin}. For all models, our method performs better than the state-of-the-art uniform quantization method. Specifically, for 3- or 4-bit quantization, our method outperforms LSQ+ \cite{lsq+} by 1.5 \% and 0.33 \% on DeiT-Tiny, respectively. We also visualize the learned bit-width allocation for the MSA layers and GELU \cite{gelu} activation.

Our main contributions can be summarized as follows:
\begin{itemize}
    \item We analyse the quantization robustness of all components in the transformer layers. In this analysis, we find that MSA and GELU are sensitive to quantization. This analysis shows the key aspect for ViT quantization and provide some insights for the future work.
    \item Based on this analysis, we propose Q-ViT, a fully differentiable quantization method, to automatically learn the optimal bit-width allocation for different components. In Q-ViT, we introduce head-wise bit-width and switchable scale to boost the performance of the quantized model. To our knowledge, it is the first work to push the limit of ViT quantization to 3-bit.
    \item Extensive experiments on different ViT models are conducted to demonstrate the effectiveness of our proposed method. Our method outperforms most quantization methods in various settings. We also do quantity and quality experiments to show the effectiveness of our proposed head-wise bit-width and switchable scale.
\end{itemize}

\section{Related Work}
\paragraph{Vision Transformer:} In computer vision, vision transformer (ViT) has become an important architecture and improved performance of many vision tasks, such as image classification \cite{deit, vit, swin}, object detection \cite{detr} and semantic segmentation \cite{swin}. It takes a sequence of flattened image patches as input, uses self-attention mechanism to develop long-range relationship, and outputs powerful image features for vision tasks. As the self-attention mechanism in the vision transformer is applied to all of the image patches, the computational cost of ViT is quadratic with the length of the image patches. Thus, through it has powerful representation, ViT is troubled with heavy computational cost. To reduce the computational cost, many new works \cite{reuse, dynamicvit, iared, tokenlearner, combiner} are proposed. \cite{iared} dynamically drops less informative patches in the original input sequence to reduce computational cost. \cite{reuse} reuses attention scores computed in one layer in multiple subsequent layers, reducing both compute and memory usage. However, there is few work about ViT quantization, though it is an alternative way to solve this problem.

\paragraph{Post-Training Quantization (PTQ):}


\paragraph{Quantization-Aware Training (QAT):} Different from  post-training quantization, quantization-aware training \cite{Gupta2015, jacob2018cvpr, louizos2018relaxed, pact, qil, tqt, lsq, zhao2019linear, lsq+} models quantization as an optimization problem, and learns the optimal quantized weights with the supervision of model performance.  LSQ \cite{lsq} learns quantization scales for each layer and reports no accuracy drop for 3-bit quantization. EdMIPS \cite{edmips} utilizes NAS to search a mixed-precision network to boost the performance of quantized networks. Compared with PTQ, QAT can obtain full precision accuracy in low-bit quantization. In this way, QAT is able to reduce computation and memory costs greatly. However, there is no QAT work for ViTs. Unlike kernel-wise or layer-wise QAT in CNNs, we design a head-wise QAT, which is fit for the multi-head attention mechanism in ViTs.

\paragraph{Transformer Quantization in Natural Language Processing (NLP) :} unlike ViT quantization, there are some works \cite{q8bert, qbert} about the quantization of transformer architecture in NLP. Q8BERT \cite{q8bert} quantizes the word embedding layers and fully connected layers in BERT to 8 bits with no severe performance drop on the General Language Understanding Evaluation (GLUE) \cite{glue} dataset. Q-BERT \cite{qbert} applies quantization to all layers in the encoder and uses a Hessian-based mixed-precision method to achieve low-bit quantization. However, these works are designed for NLP tasks and don't take advantage of the characteristic of vision transformer. Our method aims at leveraging the multi-head mechanism to achieve higher performance and propose a differentiable quantization specially designed for ViT.

\section{Method}
In this section, we firstly introduce the transformer architecture briefly. Secondly, we analyze the quantization robustness of different architecture components in the transformer layers. Finally, based on these analyses, we introduce a novel differentiable low-bit QAT method with switchable scale for ViT: Q-ViT. 

\subsection{Transformer Architecture}
In vision transformer (ViT), the transformer layer is the elementary unit. It takes a sequence of image patches as inputs, uses the self-attention mechanism to develop a long-range relationship, and outputs a new feature sequence. The transformer layer can be formulated as 
\begin{align}
    \hat X & = \text{LayerNorm}(X+\text{MSA}(X)) \\
    Y & = \text{LayerNorm}(\hat X+\text{MLP}(\hat X))
\end{align}
where  $X \in \mathbb R^{n \times d}$ is the input sequence, $n$ is the sequence length, $d$ is the feature dimension. $\hat X \in \mathbb R^{n \times d}$ is the intermediate feature sequence. $Y \in \mathbb R^{n \times d}$ is the output feature of a transformer layer. LayerNorm is the layer normalization \cite{ln}. MSA and MLP are the multi-head self-attention module and the fully connected feed-forward module, respectively. Different from conventional networks, there are two special components in transformer layers: MSA and MLP with GELU.

\paragraph{Multi-head Self-Attention:} MSA consists of $h$ heads, and each head conducts the scaled dot-product self-attention independently. For $i$-th head, the input $X$ is firstly projected into the query, key and value embeddings with learned parameter $W_i^Q, W_i^K, W_i^V$. 
\begin{equation}
Q_i = XW_i^Q, K_i = XW_i^K, V_i = XW_i^V
\end{equation}
where $W_i^Q , W_i^K, W_i^V \in \mathbb R^{d \times d_h}$. \\
Then the output of $i$-th head is computed as:
\begin{equation}
head_i = \text{softmax}(\frac{Q_i K_i^T }{\sqrt{d_k}})V_i
\end{equation}
With outputs from all heads, the MSA projects their concatenation with learned parameter $W^O$ to obtain its final output:
\begin{equation}\label{eq:proj}
    \text{MSA}(X) = \text{Concat}(head_1, \ldots, head_h) W^O
\end{equation}
where $W^O \in \mathbb R^{h d_h \times d}$.

\paragraph{MLP with GELU:} Unlike MLP layers in CNNs, the MLP layers in ViT have a different activation layer: Gaussian Error Linear Units (GELU) \cite{gelu}. It can be formulated as
\begin{equation}
\text{MLP}(\hat X) = \text{GELU}(\hat X W_1+b_1)W_2 + b_2. \label{eq:ffn}
\end{equation}
where $W_1 \in  \mathbb{R}^{d \times d_m}, b_1 \in \mathbb{R}^{d_{m}}$ and $W_2 \in \mathbb{R}^{d_{m}\times d},b_2\in \mathbb{R}^{d}$, respectively. $d_{m}$ is the hidden embedding dimension in MLP.  


\subsection{Quantization Robustness} \label{quantization-robustness-analysis}
For above special architectures in transformer layers, we analyze their quantization robustness. As is shown in ~\cref{fig:sensitivity}, we conduct ablation studies to estimate their quantization sensitivity.

First, we quantize the GELU activation layers and the fully-connected layers in MLP individually. As is shown in \cref{fig:sensitivity-a}, GELU dominates the quantization sensitivity of MLP. When not quantizing GELU, the quantized model approaches the performance of the full precision model, indicating that the quantization error introduced by other components in MLP is negligible. Thus, quantization for GELU activation requires higher bit-widths to compensate its quantization error.

Second, we analyze the quantization robustness for different heads in a specific MSA layer. We start by maintaining a uniform 8-bit DeiT-Small model quantized using a typical MSE-based PTQ approach. Then each time we set one of the heads in its first MSA layer to 2-bit and measure the accuracy drop. The result is shown in \cref{fig:sensitivity-b}. Larger performance drop indicates poorer quantization robustness. The result shows some of the heads are robust and can be quantized to low-bit representations with mild performance drop, while other heads are more crucial and need to be assigned with higher bit-widths. For example, quantizing the 2nd head to 2-bit leads to severer performance drop than the 3rd head (-1.1\% vs. -0.22\%).




\subsection{Q-ViT}
Considering the different quantization robustness of transformer components, we propose a fully differentiable QAT scheme: Q-ViT.  In Q-ViT, we introduce two learnable parameters: quantization scale $\alpha$ and float bit-width $\tilde b$. In standard QAT, bit-width $b$ is discrete. Thus, in Q-ViT, the discrete bit-width is computed as follows: 
\begin{equation}
b = \lfloor \text{clamp}( \tilde b, b_{\text{min}}, b_{\text{max}}) \rceil
\end{equation}
where $b_{\text{min}}$ is the minimum of bit-width and $b_{\text{max}}$ is the maximum.  $\text{clamp}(z, r_1, r_2)$ returns $z$ with values below $r_1$ set to $r_1$ and values above $r_2$ set to $r_2$. $\lfloor z \rceil$ rounds $z$ to the nearest integer. In our experiments, we set $b_{\text{min}}=2$ and $b_{\text{max}}=8$. 

With the quantization scale $\alpha$ and discrete bit-width $b$, we can quantize weights and activations. For a float data $x$, its quantized representation $\hat x$ is computed as:
\begin{equation}\label{eq:fp1}
\hat x = \alpha \cdot \lfloor \text{clamp}(\frac{x}{\alpha}, -q_{\text{min}}, q_{\text{max}}) \rceil
\end{equation}
where $q_{\text{max}}, q_{\text{min}}$ are the number of positive and negative quantization levels, respectively. Providing quantization bit-width $b$, unsigned integers have $q_{\text{min}}=0$ and $q_{\text{max}}=2^b-1$ and signed integers have $q_{\text{min}}=2^{b-1}$ and $q_{\text{max}}=2^{b-1}-1$. In this way, all the weights and activations are quantized by $\alpha$ and $\tilde b$. Given that the weights and activations are optimized based on the training loss, these two parameter $\alpha$ and $\tilde b$ also involve in the network training. Following LSQ \cite{lsq} and DQ \cite{dq}, we adopt scaled step size gradient and Straight-Through-Estimator (STE) \cite{bengio2013estimating} to update these two parameters. 

\begin{figure*}[t]
  \centering
  \includegraphics[width=1\linewidth]{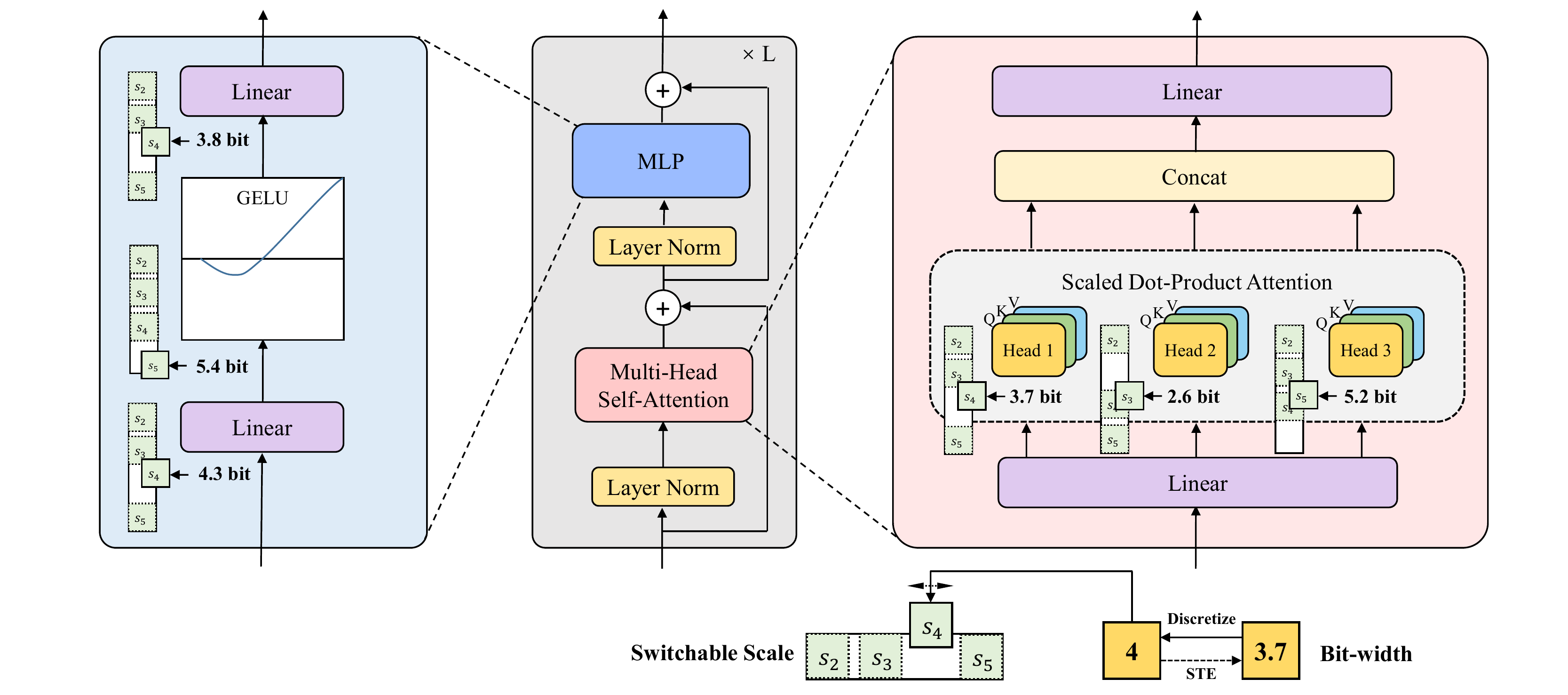}
  \caption{Illustrations for the crucial mechanism of the proposed Q-ViT. For simplicity the figure only shows a single scale set for a specific attention head in MSA, while actually all embeddings including $Q$, $K$, $V$ and attention scores have individual scale sets. And other layers such as the linear projection in MSA, the patch embedding layer and the last classification layer also are quantized using the same method but not shown on the figure.}
  \label{fig:main}
\end{figure*}

However, although the above differentiable method can allocate bit-widths to a specific component adaptively based on its contribution to the training loss, there are two shortcomings for this method: 1) this quantization method neglects the architectural characteristic of ViTs; 2) there exists an instability problem in the joint training of quantization scales and bit-widths in this method. In order to resolve these disadvantages, we introduce head-wise bit-width and switchable scale, as shown in \cref{fig:main}.

\paragraph{Head-wise Bit-width:} Unlike CNN, there is a special architecture in ViT: MSA. As we discuss in \cref{quantization-robustness-analysis}, the quantization sensitivity varies largely among different heads. Thus, different heads should be allocated with different bit-widths based on their quantization robustness. Inspired by this discovery, we propose head-wise bit-width for ViT. Specifically, each head of the embeddings in MSA maintains an independent bit-width, including $Q_i$, $K_i$, $V_i$, attention scores $A_i$ and the attention outputs $head_i$. Taking $head_i$ for example, the quantizer can be formulated as:
\begin{equation}\label{eq:quantizer}
\widehat{head_i} = \mathcal Q(head_i, \alpha_i, b_i)
\end{equation}
where $\mathcal Q(\cdot, \cdot, \cdot)$ is the proposed quantizer and $\widehat {head_i}$ stands for the quantized representation for $head_i$. $\alpha_i, b_i$ are the quantization scale and bit-width for the $i$-th head, respectively.
In the meantime, the projection weights for the specific head, namely, $W_i^Q$, $W_i^K$, $W_i^V$, $W_i^O$, also use head-wise bit-width. 
Combined with \cref{eq:quantizer}, \cref{eq:proj} can be reformulated as:
\begin{equation}\label{eq:separate}
    \widehat{MSA}(X) = \sum_{i=0}^{h} \widehat{head_i} \cdot \widehat{W_i^O},
\end{equation}
where $\widehat{W_i^O} \in \mathbb{R}^{d_h \times d}$ is the quantized representation for $W_i^O$. Note that $W_i^O$ is a part of $W^O = \text{Concat}(W_1^O, W_2^O, \cdots, W_h^O)$.

Compared with existing kernel-wise mixed-precision quantization \cite{fracbits, autoq} for CNN, head-wise bit-width for ViT has following strengths: In CNN, kernel-wise mixed-precision can only be applied to model weights. On the one hand, the features are unable to utilize channel-wise mixed-precision because all channels in input features are involved in the computation for one output channel, and can not be separated like ViT as \cref{eq:separate}. On the other hand, quantizing the input features with different bit-widths for different weight kernels would also introduce large computation overhead, as is discussed in \cite{fracbits, autoq}. In contrast, in the self-attention mechanism of ViT, the dimension for head does not participate in the matrix multiplication and each head performs the self-attention calculation individually. Thus, leveraging head-wise bit-width doesn't require expensive computation overhead.

\paragraph{Switchable Scale:} In our preliminary experiments, we find that the quantization scale $\alpha$ is unstable during the training. In other words, its value changes dramatically in the training and leads to poor model performance. After rethinking the relation between scales and bit-widths, we attribute this to the shift of bit-width. In specific, the relation between the scale $\alpha$ and the bit-width $b$ is defined by
\begin{equation}
\label{relation}
\alpha = \frac{x_\text{max}-x_\text{min}}{2^b}
\end{equation}
where $x_\text{max}, x_\text{min}$ is the maximum and minimum quantization range for x, respectively. Assuming $x_\text{max}-x_\text{min}$ is fixed, the optimal solution for $\alpha$ would change exponentially with the variation of bit-width $b$.

Inspired by switchable batch normalization in \cite{slimmable}, we propose a novel technique named switchable scale to solve the unstable training problem. In details, we expand $\alpha$ into a learnable vector $\boldsymbol{\alpha}$. Each $\boldsymbol{\alpha_i}$ is responsible for a candidate bit from $b \in \{2, 3, \cdots, 8\}$.  This enables independent learning for the corresponding scale when the bit-width searches among different bits. In this way, there exists no dramatic shift problem, leading to a stable optimization for the joint training of scales and bit-widths. Note that switchable scale does not introduce any extra computational cost in both the training and the inference stage because only the scale w.r.t the corresponding bit will participate in the prediction. As is further shown in \cref{tab.ablations.sqs}, switchable scale provides a significant performance boost for Q-ViT.
\begin{table*}[t]
  \centering
  \begin{tabular}{cccccc}
    \toprule
    \textbf{Model} & \textbf{Method}& \textbf{W-bit}& \textbf{A-bit}&\textbf{{BitOPs (G)}}&\textbf{{Top-1}}\\
    \midrule
    \multirow{5}{*}{DeiT-T} & Baseline & float & float & -- & 72.86 \\
    \cline{2-6}
     & LSQ+\cite{lsq+} & 4 & 4 & 21.5 & 72.46 \\
     & Q-ViT (ours) & learned & learned & 21.4 & \textbf{72.79} \\
    \cline{2-6}
     & LSQ+ & 3 & 3 & 12.9 & 68.09 \\
     & Q-ViT (ours) & learned & learned & 13.0 & \textbf{69.62} \\
    \midrule
    \multirow{5}{*}{DeiT-S} & Baseline & float & float & -- & 79.92 \\
    \cline{2-6}
     & LSQ+ & 4 & 4 & 76.4 & 79.66 \\
     & Q-ViT (ours) & learned & learned & 77.9 & \textbf{80.11}\\
    \cline{2-6}
     & LSQ+ & 3 & 3 & 44.6 & 77.76 \\
     & Q-ViT (ours) & learned & learned & 45.7 & \textbf{78.08} \\
    \midrule
    \multirow{5}{*}{Swin-T} & Baseline & float & float & -- & 80.90 \\
    \cline{2-6}
     & LSQ+ & 4 & 4 & 72.6 & 80.47 \\
     & Q-ViT (ours) & learned & learned & \ \ \ \ 70.1 $\downdownarrows$ & \textbf{80.59} \\
    \cline{2-6}
     & LSQ+ & 3 & 3 & 41.3 & 78.96\\
     & Q-ViT (ours) & learned & learned & \ \ \ 40.8 $\downarrow$ & \textbf{79.45}  \\
    \bottomrule
  \end{tabular}
  \caption{Compared with uniform quantization on ImageNet. $ \downarrow$ \& $\downdownarrows$ stands for observing an noticable \& considerable reduction on BitOPs compared with uniform quantization using our method.}
  \label{tab:sota}
\end{table*}
\paragraph{Complexity Constraint:} Like existing differentiable mix-precision quantization \cite{dq, fracbits, edmips}, Q-ViT also has a complexity constraint. In previous works, the complexity constraint is miscellaneous, such as memory footprints \cite{dq}, model size \cite{dq}, BitOPs \cite{oneshotnas, fracbits}. In Q-ViT, we adopt BitOPs as the complexity constraint because it is hardware-agnostic and can be calculated with model definition and bit-width allocation directly. With this constraint, the optimization for Q-ViT under the BitOPs constraint $c$ can be formulated as:
\begin{equation}
\mathop{\min}_{W, S, B} E_{X \sim D }[\mathcal F(X, W, S, B)] \ \ \ \text{ s.t. } \mathcal C(\mathcal F, B) \le c
\end{equation}
where $X$ is the input image, sampled from data distribution $D$. $W, S, B$ are the weight parameters, the quantization scale set and the bit-width set of the quantized model $\mathcal F$, respectively. $\mathcal C(\cdot, \cdot)$ is the BitOPs of the quantized model $\mathcal F$ given bit-width set $B$.

For this constrained optimization problem, we adopt the penalty method \cite{penalty} to turn the constraint into a regularization term:
\begin{equation}
\mathop{\min}_{W, S, B} E_{X \sim D }[\mathcal F(X, W, S, B)] + \eta [H(\mathcal C(\mathcal F, B) - c)]^2
\end{equation}
where $\eta$ is the regularization coefficient. $H(\cdot)$ is as follows:
\begin{equation}
H(z) = 
\begin{cases}
z & \text{if $z > 0$} \\
0 & \text{otherwise}
\end{cases}
\end{equation}
More details are shown in \cref{tab.ablations.total}.

\section{Experiments}\label{sec:exp}
We evaluate the effectiveness of our Q-ViT on ViT-variant models, such as DeiT \cite{deit} and Swin Transformer \cite{swin}. In the following experiments, we report Top-1 accuracy on ImageNet \cite{imagenet} and BitOPs of models. Note that for simplicity, we use the term $N$-bit constraint in the following sections, inferring to the Q-ViT model is trained under the same BitOPs constraint with a uniform N-bit model.

\subsection{Implementation Details} In our experiments, we initialize the weights of quantized model with the corresponding pretrained float model. For the initialization of quantization parameters, such as quantization scales and bit-widths, we set the initial bit-widths to be $N+1$ when the complexity constraint is $N$-bit except that the patch embedding (first) layer and the classification (last) layer are 8-bit. Note that unlike previous works \cite{fracbits, etnnq}, the quantization parameters of first and last layer are optimized and are not fixed during the training. We initialize the all scales in the switchable scale vectors using a typical MSE-based approach.

The quantized model is trained for 300 epochs with batchsize 512 and the base learning rate 2e-4. We do not use warm-up scheme and Pytorch AMP \footnote{https://pytorch.org/docs/1.7.1/amp.html?highlight=amp}. For AMP, we find it causes a numerical instability for gradient when AMP is used in our experiments. Disabling AMP has no effect on the performance in our experiments. We also adopt a two-stage training pipeline as previous work \cite{fracbits} does. This two-stage pipeline is divided into the searching stage and the diving stage. In the searching stage, bit-widths and model parameters are optimized simultaneously. The quantized model will search the optimal bit-width allocation. In the diving stage, bit-widths are fixed and only model parameters are trained. In this paper, we introduce a hyper-parameter $\sigma \in [0,1]$ to control the ratio of searching stage during the whole training. For all the experiments, we set $\sigma = 0.9$.  Other training settings follow DeiT \cite{deit} or Swin Transformer \cite{swin}. 

\begin{figure*}
    \centering
    \begin{subfigure}{0.24\linewidth}
        \includegraphics[width=1\linewidth]{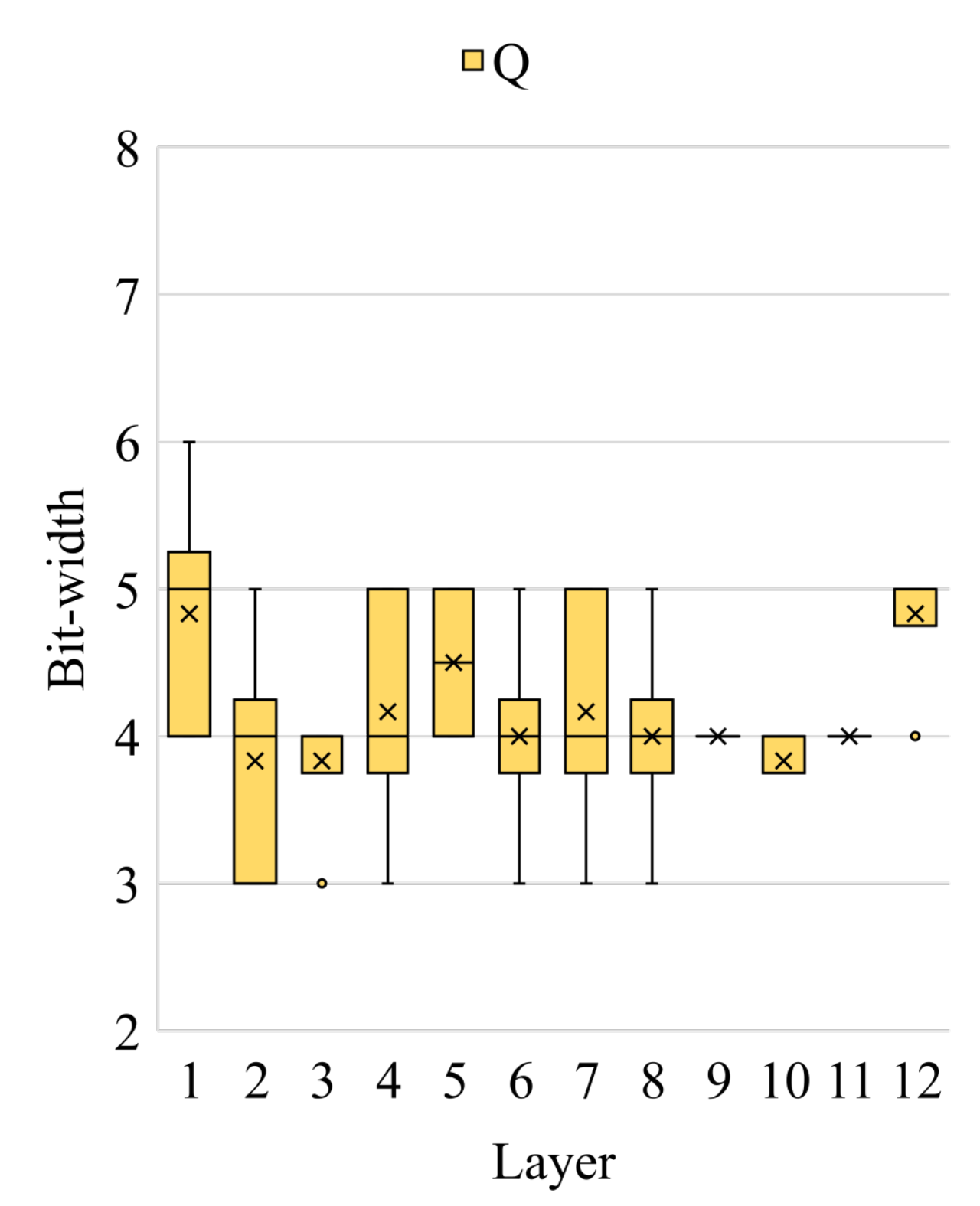}
        \caption{}
        \label{fig:bit:q}
    \end{subfigure}
    \hfill
    \begin{subfigure}{0.24\linewidth}
        \includegraphics[width=1\linewidth]{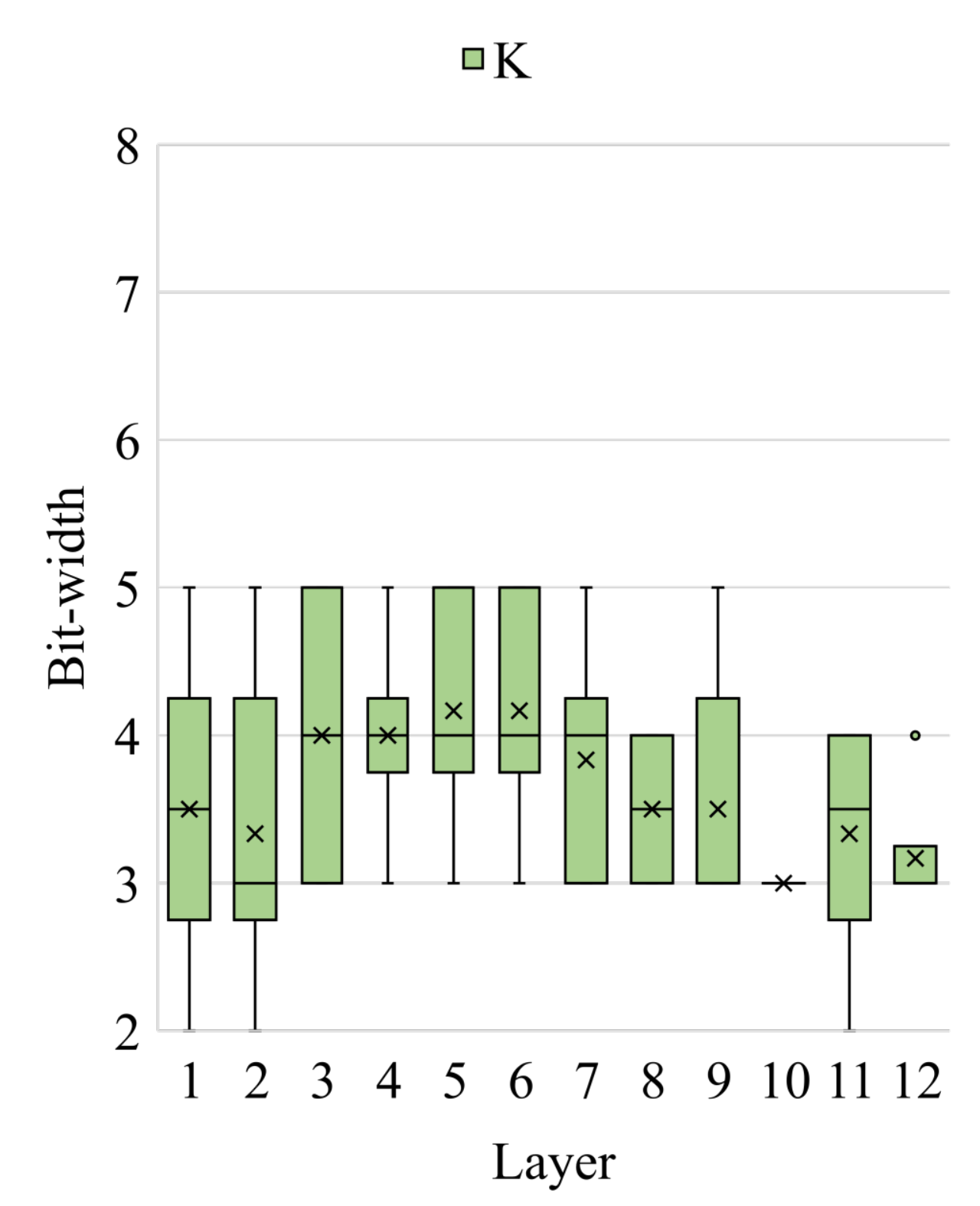}
        \caption{}
        \label{fig:bit:k}
    \end{subfigure}
    \begin{subfigure}{0.24\linewidth}
        \includegraphics[width=1\linewidth]{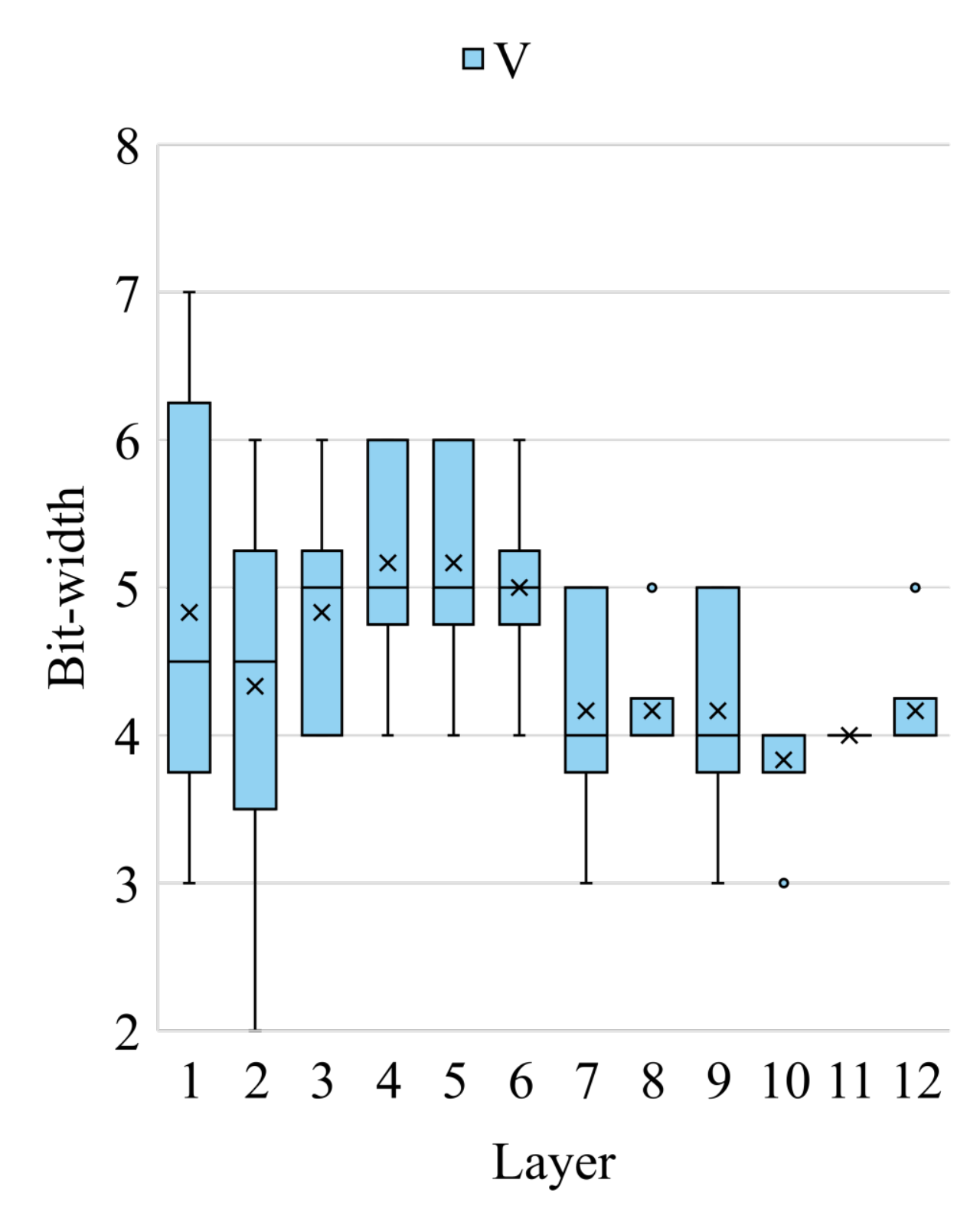}
        \caption{}
        \label{fig:bit:v}
    \end{subfigure}
    \hfill
    \begin{subfigure}{0.24\linewidth}
        \includegraphics[width=1\linewidth]{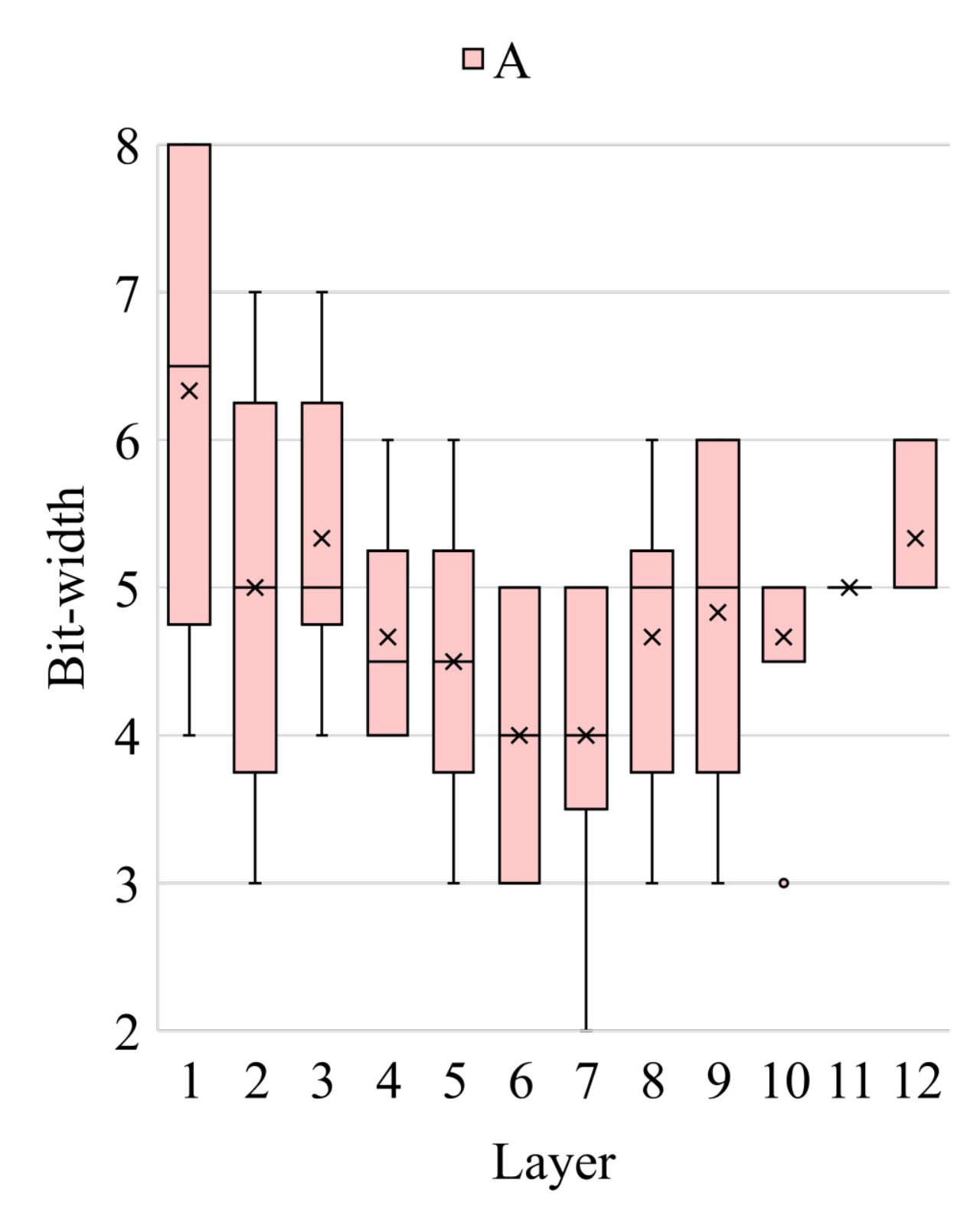}
        \caption{}
        \label{fig:bit:a}
    \end{subfigure}
    \caption{Learned bit-width allocation for different attention heads of Q, K, V and attention scores in a 4-bit constrained DeiT-Small model, visualized using box-plot. The colored boxes stand for the learned bit-width span of different heads. The intermediate short dash is for the median bit-width, while the short dashes on both ends are for the minimum and maximum bit-width. "$\circ$" is for the outlier and "$\times$" is for the average bit-width.}
    \label{fig:bit:qkva}
\end{figure*}
\begin{figure}[t]
    \centering
    \includegraphics[width=0.8\linewidth]{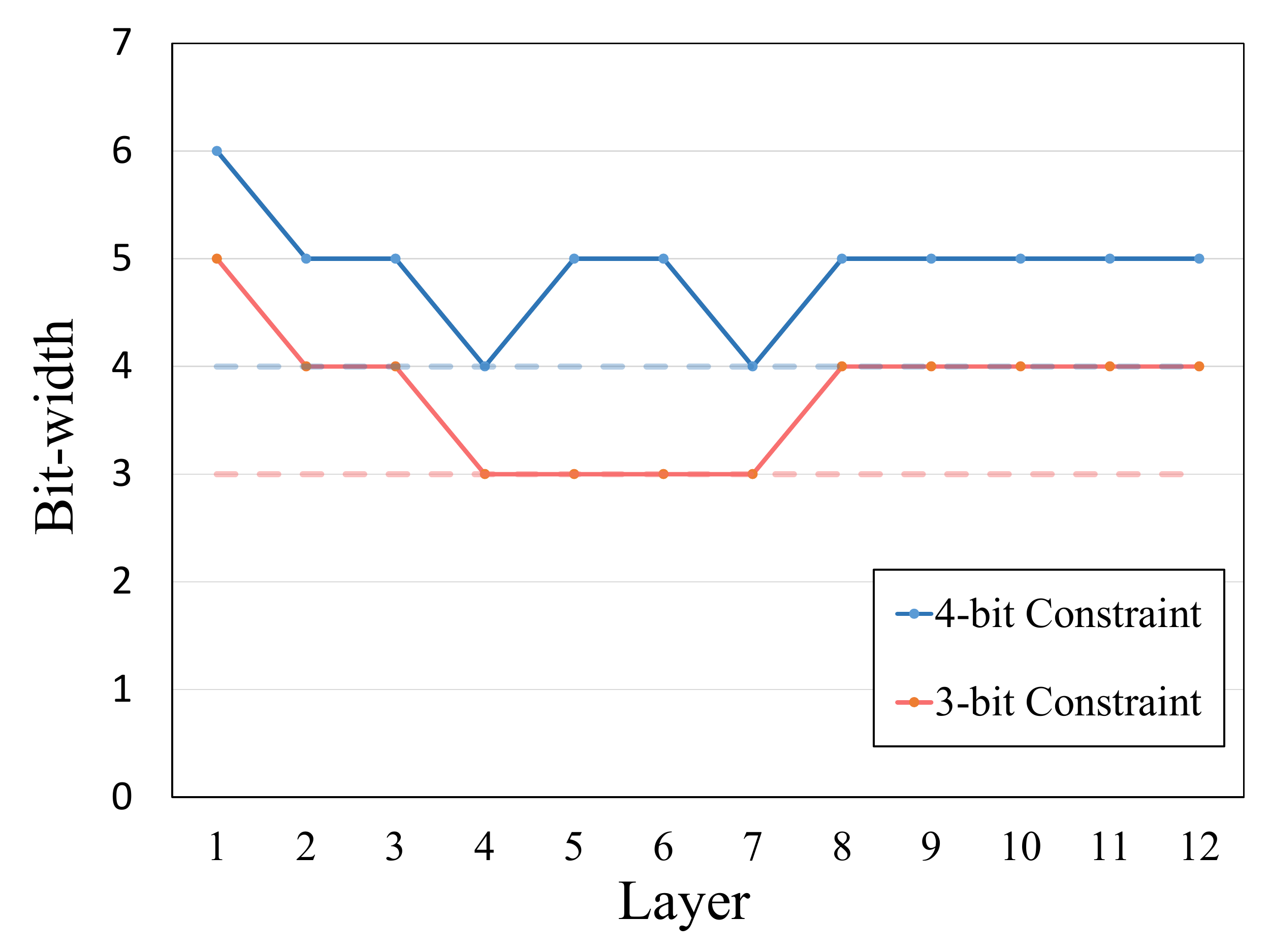}
    \caption{Learned bit-width allocation for GELU in DeiT-Tiny.}
    \label{fig:bit:gelu}
\end{figure}
\subsection{Main Results} Given that there is no QAT work on ViT, we implement LSQ+ \cite{lsq+} on ViT and compare our method with it. For LSQ+, we replace Gaussian-based initialization method with MSE-based initialization method. As is shown in \cref{tab:sota}, Q-ViT outperforms LSQ+ on different ViT-variants and complexity settings. In particular, Q-ViT surpasses 3-bit LSQ+ by a large margin of 1.5\% on DeiT-Tiny. It also outperforms LSQ+ by 0.5\% on Swin-Tiny with noticeable BitOPs reduction. These results prove the effectiveness of Q-ViT.  In addition, we find that DeiT-Tiny is relatively more sensitive to quantization than DeiT-Small and Swin-Tiny. For uniform 3-bit quantization, the performance drop for DeiT-Small and Swin-Tiny is about 2\%, while for DeiT-Tiny, the drop is nearly 5\%. We conjecture the reason is that DeiT-Small and Swin-Tiny have more abundant parameters, which makes them more robust to quantization. This also explains why DeiT-Tiny benefits more from our method.

\subsection{Learned Bit-width Allocation} For a further understanding about how Q-ViT works and its mechanism behind the performance improvements, we visualize the learned bit-width allocation of different components in ViT.
\paragraph{First \& Last layers:} As we mentioned before, different from previous work \cite{fracbits, etnnq}, we enable bit-width learning for the patch embedding (first) layer and the classification layer (last) layer in ViT. The standard practice in quantization is that the first and last layer in a deep neural network are allocated with high bit-width, \eg 8-bit. When the bit-widths of these layers are learned, the final bit-widths of these layers are shown in ~\cref{tab:first-last}. We find that the bit-widths of these layers are high in different ViT-variants. This demonstrates these two layers are important and our method manages to assign high bit-widths for them, which is consistent with standard practice. Moreover, we also find that the bit-width of the first layer sometimes reaches 4-bit while the bit-width of the last layer is always about 8-bit in ViT. This observation shows that the last layer is more important than the first layer. In addition, the phenomenon that our method assigns lower bit-widths to weights than activations shows that activations are more important than weights in these layers. 


\begin{table}
    \begin{tabular}{cccc}
        \toprule
        \textbf{Model} & \textbf{Constraint}& \textbf{Patch Embed}&\textbf{{Classification}}\\
        \multirow{2}{*}{DeiT-T} & 4-bit & 5 / 7 & 8 / 8 \\
         & 3-bit & 4 / 6 & 7 / 8 \\
        \midrule
        \multirow{2}{*}{DeiT-S} & 4-bit & 4 / 7 & 8 / 8  \\
         & 3-bit & 6 / 8 & 8 / 8 \\
        \midrule
        \multirow{2}{*}{Swin-T} & 4-bit & 8 / 8 & 8 / 8 \\
         & 3-bit & 6 / 8 & 7 / 8 \\
        \bottomrule
    \end{tabular}
    \caption{Bit-width allocation of the patch embedding layer and the classification layer learned in different models and constraints. The results are given in the format W / A, for bit-widths of weight and activation, respectively.}
    \label{tab:first-last}
\end{table}

\begin{table*}[t]\vspace{0mm}
\footnotesize
\centering

\subfloat[\textbf{Head-wise Bit-width:} The effectiveness of head-wise bit-width.\label{tab.ablations.hwbw}]{
\begin{minipage}{.25\textwidth}
\centering
\tablestyle{5pt}{1.05}\setlength{\tabcolsep}{1.mm}\
    \begin{tabular}{ccccc}
        \toprule
        \textbf{Model} & \textbf{Constraint} & \textbf{HW} & \textbf{{BitOPs(G)}} & \textbf{{Top-1}}\\
        \midrule
        \multirow{2}{*}{DeiT-T}& 3-bit & $\times$ & 13.0 & 69.15 \\
        & 3-bit & $\checkmark$ & 13.1 & \textbf{69.62} \\
        \midrule
        \multirow{2}{*}{DeiT-S}& 3-bit & $\times$ & 44.3  & 77.77 \\
        & 3-bit & $\checkmark$ & 45.7  & \textbf{78.08} \\
        \midrule
        \multirow{2}{*}{DeiT-S}& 4-bit & $\times$ & 75.5  & 79.69 \\
        & 4-bit & $\checkmark$ & 77.9  & \textbf{80.11} \\
        \bottomrule
    \end{tabular}
\end{minipage}
}\hspace{1.1cm}
\subfloat[\textbf{Switchable Quantization Scale:} The effectiveness of switchable quantization scale.\label{tab.ablations.sqs}]{
\begin{minipage}{.3\textwidth}
\tablestyle{5pt}{1.05}\setlength{\tabcolsep}{1.mm}\
  \begin{tabular}{ccccc}
    \toprule
    \textbf{Model} & \textbf{Constraint} & \textbf{SS} & \textbf{{BitOPs(G)}} & \textbf{{Top-1}}\\
    \midrule
    \multirow{3}{*}{DeiT-T} & Uniform & -- & 21.5 & 72.46 \\
     & 4-bit & $\times$ &  21.4 & 71.62 \\
     & 4-bit & $\checkmark$ &  21.9 & \textbf{72.97} \\
    \midrule
    \multirow{3}{*}{DeiT-T} & Uniform & -- & 12.9 & 68.09 \\
     & 3-bit & $\times$  & 12.8 & 67.50 \\
     & 3-bit & $\checkmark$ & 13.0 & \textbf{69.15} \\
    \bottomrule
    \end{tabular}
    \vspace{0.09cm}
\end{minipage}
}\hspace{.1cm}
\subfloat[\textbf{Hyper-parameter $\sigma$:} The effect of the hyperparameter $\sigma$ on quantized model. \label{tab.ablations.hyperparameter}]{
\begin{minipage}{.25\textwidth}
\tablestyle{5pt}{1.05}\setlength{\tabcolsep}{1.mm}\
    \begin{tabular}{ccccc}
    \toprule
    \textbf{Model} & \textbf{Constraint} & \textbf{$\sigma$} & \textbf{{BitOPs(G)}} & \textbf{{Top-1}}\\
    \midrule
    \multirow{4}{*}{DeiT-T} & 4-bit & 0.5 & 21.8 & 69.54 \\
     & 4-bit & 0.7 & 22.1 & 71.14 \\
     & 4-bit & 0.9 & 21.9 & 71.05 \\
     & 4-bit & 1 & 22.2 & 69.80 \\
    \bottomrule
    \end{tabular}
    \vspace{0.5cm}
\end{minipage}
}
\caption{\textbf{Ablations.} All ablation experiments are conducted on ImageNet data.}\label{tab.ablations.total}
\end{table*}

\paragraph{Bit-width Allocation in MSA:} In Q-ViT, we apply head-wise bit-width in MSA, which learns different bit-widths for queries ($Q$), keys ($K$), values ($V$) and attention scores ($A$). For these modules, the learned bit-widths in 4-bit DeiT-Small are shown in ~\cref{fig:bit:qkva}.  For $Q$ and $K$, they prefer to learn a lower bit-width, while $V$ and $A$ learn a relatively higher bit-width. This demonstrates that $V$ and $A$ are more important than $Q$ and $K$ and have greater contribution to performance. In all the modules, the bit-widths for different heads in early layers are diverse. In contrast, the bit-widths in last layers are similar, i.e. most heads learn the same bit-width. We speculate this is caused by the over-smoothness phenomenon of ViT discovered in \cite{gong2021improve}, i.e. the features have high similarity in deep layers of ViT. This might explain the similar bit-width allocation for deep layers in Q-ViT, whereas low feature similarity in early layers contributes to the diversity of the learned bit-widths. In addition, different learned bit-widths in ViT show that our head-wise bit-with in MSA is essential and specific for ViT. 


\paragraph{Bit-width Allocation in GELU:} In \cref{quantization-robustness-analysis}, our experiment shows that GELU is sensitive to quantization due to its long-tailed distribution. Given this, Q-ViT is expected to assign high bit-widths for GELU. As is shown in \cref{fig:bit:gelu}, we find that GELU learns high bit-widths compared with the bit-width constraint. For example, in the 3-bit constraint, most GELUs prefer to have high bit-width allocation ($\geq 4$ bit). Moreover, the GELUs have relatively higher bit-widths in the early and deep layers than in the middle layers. These observations are consistent with high importance of these layers.

\subsection{Ablation Studies}
\paragraph{Head-wise Bit-width:} To show the effectiveness of head-wise bit-width, we conduct experiments to show the performance improvement from this mechanism. As shown in ~\cref{tab.ablations.hwbw}, head-wise bit-width boosts the performance by ~0.5\% in different ViT-variants. 
\paragraph{Switchable Scale:} \cref{tab.ablations.sqs} shows the switchable scale technique that we proposed is crucial in Q-ViT. Without switchable scale, Q-ViT suffers from a severe convergence problem for both 3-bit and 4-bit BitOPs constraint, performing even worse than the uniform quantization. Enabling switchable scale provides a performance boost of about 1.5\%.
\paragraph{Hyper-parameters $\eta$ and $\sigma$:} The regularization coefficient $\eta$ affects the performance and BitOPs of the quantized model. In our experiments, we find $\eta=0.1$ gives a good trade-off between accuracy and BitOPs. When $\eta$ is set to 0.01, the performance of the quantized model is better than one under $\eta=0.1$, but the BitOPs goes beyond the given BitOPs constraint. Also, when $\eta=1$, the quantized model satisfies BitOPs constraint but has inferior performance.  

The ratio of searching stage $\sigma \in [0,1]$ also affects the training of the quantized model. As is shown in ~\cref{tab.ablations.hyperparameter}, we find $\sigma$ mainly affects the performance of the quantized model in two ways. If $\sigma$ is too small, \eg $\sigma$ = 0.5, the bit-width allocation ends up with a greedy solution and is inferior in performance. And if $\sigma$ is too large, like to the extreme, $\sigma$ = 1, which means Q-ViT are trained with bit-widths learned all the time, convergence problem for the model parameters are observed due to the frequent alterations of the bit-width allocation. When $\sigma$ is properly set, like 0.7 or 0.9, optimization for the bit-width allocation and model parameters are well-balanced, achieving better performance.
\section{Conclusion}
In this work, we analyze the quantization robustness of all components in the transformer layers. This demonstrates the quantization characteristics of ViT and provides some insights for ViT quantization.  Based on this analysis, we propose Q-ViT, a fully differentiable quantization method, which consists two features: head-wise bit-width and switchable scale. To our knowledge, it is the first work to push the limit of ViT quantization to 3-bit. Extensive experiments on different ViT models are conducted to demonstrate the effectiveness of our proposed method. 


\clearpage
{\small
\bibliographystyle{ieee_fullname}
\bibliography{cite}
}

\end{document}